\pdfoutput=1

\documentclass[11pt]{article}

\usepackage[preprint]{acl}

\usepackage{times}
\usepackage{latexsym}
\usepackage{enumitem}
\usepackage{hyperref}
\usepackage{adjustbox}

\usepackage{lipsum}
\usepackage{url}
\usepackage{graphicx}
\usepackage{color}
\usepackage{booktabs}
\usepackage{caption}
\usepackage{longtable} 
\usepackage{afterpage}
\usepackage{booktabs} 
\usepackage{amsmath}

\usepackage{longtable} 
\usepackage{geometry} 
\usepackage{balance} 
\usepackage[T1]{fontenc}

\usepackage[utf8]{inputenc}

\usepackage{microtype}

\usepackage{inconsolata}

\title{Real World Conversational Entity Linking Requires \\ More Than Zero-Shots}

\author{
  \textbf{Mohanna Hoveyda\textsuperscript{1}}\quad
  \textbf{Arjen P. de Vries\textsuperscript{1}}\quad
  \textbf{Maarten de Rijke\textsuperscript{2}}\quad
    \textbf{Faegheh Hasibi\textsuperscript{1}}\\[1ex]
  \textsuperscript{1}Radboud University, The Netherlands\\
  {\fontsize{10}{12}\selectfont
    \texttt{mohanna.hoveyda@ru.nl}, \texttt{arjen.devries@ru.nl}, \texttt{faegheh.hasibi@ru.nl}
  }\\[1ex]
  \textsuperscript{2}University of Amsterdam, The Netherlands\\
  {\fontsize{10}{12}\selectfont
    \texttt{m.derijke@uva.nl}
  }
}
\begin{document}
\maketitle
\begin{abstract}
Entity linking (EL) in conversations faces notable challenges in practical applications, primarily due to the scarcity of entity-annotated conversational datasets and sparse knowledge bases (KB) containing domain-specific, long-tail entities. 
We designed targeted evaluation scenarios to measure the efficacy of EL models under resource constraints. Our evaluation employs two KBs: Fandom, exemplifying real-world EL complexities, and the widely used Wikipedia. First, we assess EL models' ability to generalize to a new unfamiliar KB using Fandom and a novel zero-shot conversational entity linking dataset that we curated based on Reddit discussions on Fandom entities. We then evaluate the adaptability of EL models to conversational settings without prior training. Our results indicate that current zero-shot EL models falter when introduced to new, domain-specific KBs without prior training, significantly dropping in performance.
Our findings reveal that previous evaluation approaches fall short of capturing real-world complexities for zero-shot EL, highlighting the necessity for new approaches to design and assess conversational EL models to adapt to limited resources. The evaluation setup and the dataset proposed in this research are made publicly available.%
\footnote{\url{https://github.com/informagi/reddit_ConEL}}
\end{abstract}

\section{Introduction}
Entity Linking (EL) is the process of detecting and resolving ambiguous mentions of entities in a given text by accurately associating them with their corresponding entries in a knowledge base \citep{kolitsas-etal-2018-end,sevgili2022neural}.
This is a pivotal step in many downstream tasks such as semantic search \citep{Gerritse:2022:EMBERT, Chatterjee:2023:BQB, Hasibi:2016:EEL}, question answering \citep{liu2023data}, and conversational search~\cite{zamani2023conversational}. 

The significance of EL particularly comes to the fore in the realm of conversational systems as it helps to enhance the accuracy and relevance of the information provided to users during a dialogue session. As these systems are becoming increasingly prevalent in various applications, their ability to ground discussions in real-world knowledge is indispensable for maintaining the integrity and usefulness of the system
\citep{ahmadvand2019concet,Cruz2023,kandpal2023large}.

Conversations possess characteristics that render common EL models suboptimal; e.g., noisy text, informal language, and entity-related information spreading through turns \citep{joko2021conversational, joko2022personal}. However, conversational EL has been less explored in prior research and is predominantly focused on techniques and benchmarks for long  documents \citep{logeswaran-etal-2019-zero} or stand-alone queries \citep{hasibi2015entity, Hasibi:2017:ELQ}. EL approaches also often assume the existence of ample training data \citep{DBLP:conf/iclr/CaoI0P21,van2020rel,piccinno2014tagme}, a similar distribution of entities in KB during training and at inference time, and a structurally/textually rich KB for training. These assumptions, however, do not usually hold in real-world EL scenarios, especially in a conversational context, making EL in practice more challenging. 


Train and deployment of EL systems in general poses several other challenges as well. Creating an entity-annotated training dataset can be prohibitively exhaustive, or the data might be unavailable due to privacy concerns \citep{sui2023selecting}. In addition, the distribution of train and test entities might differ as knowledge bases may expand with time, and new entities can be added to the KB which results in an incomplete KB at training time \cite{aydin-etal-2022-find, zhang2018entity}. Lastly, real-world KBs do not often come with dense structural/textual entity information.

As a result, zero-shot entity linking \cite{logeswaran-etal-2019-zero, bhargav2022zero} was introduced to address some of these challenges. This setup is aimed to allow disambiguating mentions of previously unseen entities by relying on pre-trained models.

In this study, we design an evaluation framework and a dataset, addressing the gap between real-world conversational EL and the existing zero-shot EL studies, showing that current zero-shot models do not adequately address practical challenges. We pose our research questions as \textbf{RQ1)} \textit{Are zero-shot EL models able to generalize effectively when introduced to a whole new KB, not included in their initial training?} \textbf{RQ2)} \textit{How much can zero-shot EL models adapt to conversational settings without prior training?}

The contributions of this paper are: 
\begin{itemize}[noitemsep,leftmargin=*]
    \item Introducing evaluation scenarios to highlight gaps in zero-shot EL research and evaluation inadequacies specifically in conversational settings.
    \item Creating a conversational dataset to demonstrate real-world EL challenges empirically and to facilitate research in addressing practical EL challenges.
    \item Demonstrating that current zero-shot EL models significantly underperform when applied to new, domain-specific KBs without prior exposure to their entities, emphasizing that zero-shot EL is yet to be effective in solving real EL tasks.

\end{itemize}

\section{Analysis Scenarios}
To assess models based on practical constraints we perform the following groups of analysis.
\subsubsection*{Generalization to Unfamiliar KB and EL task}
This set of experiments is aimed to assess how well EL models are capable of generalizing to a new KB at inference time. Given $G$ and $G^{\prime}$ as KBs, models are previously trained on G and encounter $G^{\prime}$ only at the evaluation step. Particularly selecting $G^{\prime}$ to ensure the frequency of domain-specific and long-tail entities, makes the task more challenging. 

Our definition of generalizability diverges from the definition adopted by \cite{logeswaran-etal-2019-zero} and \cite{wu-etal-2020-scalable}. In our approach, we strictly enforce that there is no overlap between the knowledge bases used for training and evaluation. Specifically, we use models that are exclusively trained on Wikipedia ($G$) and are only exposed to the Fandom knowledge base ($G^{\prime}$) during evaluation. This contrasts with the methodologies in the cited studies, where models receive training on some segments of the Fandom knowledge base before evaluation, even though these are distinct from the test segments. 

Along unfamiliar KB scenario, we also assess the generalization of the EL systems to a new setting which is conversational EL, since these models have not been previously trained on this setting, we intend to assess their generalizability to this setting as well.

\subsubsection*{\textbf{Adaptability to Conversational EL Task}}
In the second set of evaluation experiments, we examine how well EL models perform in a conversational setting. We formulate this as a zero-shot EL task since it tests the model's adaptability to a new setting (i.e., conversational), 
given that zero-shot EL models are typically trained for documents, 
and queries and not conversations.

\begin{table}
  \begin{tabular}{lcl}
  \toprule
     &\textbf{Train}&\textbf{Test}\\
    \midrule
    Conversations & 5352 & 745\\
    Threads & 8026 & 745 \\
    All utterances & 49695 & 4557 \\
    Annotations & 10263 & 965 \\
    Utterances with Annotations & 8787 & 833 \\
    Average thread length & 6.19 & 6.11 \\
  \bottomrule
\end{tabular}
  \caption{Reddit Conversational Data Statistics}
\label{tab:freq}
\end{table}

\begin{table*}[t] 
\centering
\small
\setlength{\tabcolsep}{1pt} 
\begin{tabular}{l ccc ccc ccc ccc ccc ccc}
\toprule
& \multicolumn{9}{c}{\textbf{Wikia}} & \multicolumn{9}{c}{\textbf{Reddit}} \\
\cmidrule(r){2-10}
\cmidrule{11-19}
 & \multicolumn{3}{c}{MD} & \multicolumn{3}{c}{ED} & \multicolumn{3}{c}{EL} & \multicolumn{3}{c}{MD} & \multicolumn{3}{c}{ED} & \multicolumn{3}{c}{EL} \\
\cmidrule(r){2-4}
\cmidrule(r){5-7}
\cmidrule(r){8-10}
\cmidrule(r){11-13}
\cmidrule(r){14-16}
\cmidrule{17-19}
 & P & R & F & P & R & F & P & R & F & P & R & F & P & R & F & P & R & F \\
\midrule
\textbf{Flair + BLINK} Micro &.027 & .255 & .048 & .026 & .222 & .047 & .015 & .147 & .027 & .130 & .186 & .153 & .167 & .232 & .194 & .064 & .093 & .076  \\
\textbf{Flair + BLINK} Macro & .029 & .269  & .051 & .029 & .241 & .051 & .015 & .156 & .028 & .136 & .202 & .162 & .160 & .237 & .191 & .057 & .088 & .069 \\
\midrule
\textbf{ELQ} Micro  & .034 & .205& .058 & .015 & .088 & .025 & .010 & .062 & .017 & .135 & .313 & .189  & .162 & .367 & .225 & .069 & .161 & .097  \\
\textbf{ELQ} Macro  & .036 & .223 & .062 & .019 & .117 & .033 & .013 & .081 & .022  & .123 & .285 & .171 & .142 & .323 & .197 & .057 & .134 & .080 \\
\bottomrule
\end{tabular}
\caption{Entity linking micro and macro-averaged scores on Reddit dataset using Fandom as the knowledge base MD, ED, and EL show the relevant scores for mention detection, entity disambiguation, and entity linking. The scores indicate precision (P), recall (R), and f1-score (F). Only the corresponding domain knowledge base is used for each domain at inference time.}
\label{table:reddit_results}

\end{table*}

 \section{Reddit Conversational Dataset for Zero-shot EL}
 \label{Reddit_data_desc}
We introduce the Reddit Conversational EL dataset, specifically curated for evaluating zero-shot EL methods in conversational setup and with unseen KBs, used for our analysis scenarios. 
\\
To curate this dataset we used the Convokit's Reddit corpus%
\footnote{\url{https://convokit.cornell.edu/documentation/subreddit.html}} \cite{chang-etal-2020-convokit}, which includes subreddit posts and comments until October 2018, sourced from the broader Pushshift Reddit dataset%
\footnote{\url{https://pushshift.io/}} \cite{baumgartner2020pushshift}.
Convokit offers 948,169 subreddits, among which, we only opt for the discussions around each of the 16 domains used in ZESHEL~\cite{logeswaran-etal-2019-zero} (introduced in \ref{subsec:kbs}). We extract subreddits that contain a ZESHEL domain title in their name. From each Reddit conversation, we extract its unique threads. In this context, a thread is a distinct path in a hierarchical structure of user utterances, beginning with an original post (the root) and encompassing all subsequent replies until the last reply (the leaf) \cite{zhang-etal-2020-dialogpt,henderson-etal-2019-repository}. 
To create gold mention spans along with their gold Fandom entities, we rely on instances where users include hyperlinks to the Fandom website as a way of disambiguating their mention of an entity in their utterance. Next, several preprocessing, pruning, and augmentation steps were performed:
\begin{enumerate}[noitemsep,leftmargin=*]
    \item Removed URLs, special symbols, non-English characters, and repetitive nonsensical tokens.
    \item Pruned utterances including profanity keywords (based on a publicly available profanity list \cite{harel2022dataset}) and utterances with less than 5 or more than 70 tokens
    \item Excluded annotations with nonsensical mentions (e.g; "here", "this link", "link" etc.)
    \item Augmented user annotations in cases where the exact mention text is annotated by the user in some occurrences but not others
    \item Excluded threads with less than 5 utterances and threads with no annotations
\end{enumerate}
We checked the extracted annotations for instances where the gold mention and entity were exact matches. To avoid trivial disambiguation tasks, following \cite{logeswaran-etal-2019-zero}, we ensured no more than 5\% of our threads have such annotations.
Splitting the final data to train and test sets, we relied on conversation timestamps and annotation density (details in Appendix \ref{sec:data_split}). Dataset statistics can be found in Table~\ref{tab:freq}. Samples of the dataset are included in Table \ref{tab:reddit_samples}.

\section{Experimental Setup}
\label{sec:Exp}
We detail the entity linking models, datasets and knowledge bases used in our experiments, as well as experimental details of our analysis setups.

\subsection{Entity Linking Models}
We focus on assessing two of the very few models purported to facilitate zero-shot entity linking; ELQ \cite{li-etal-2020-efficient} and BLINK \citep{wu-etal-2020-scalable}, both BERT-based models that are pre-trained on Wikipedia for EL. ELQ is based on a biencoder model and performs mention detection and entity disambiguation simultaneously in a single pass, showing promise in zero-shot QA contexts. Our analysis, however evaluates its ability to adapt to conversations. BLINK, on the other hand, specializes in entity disambiguation, requiring either predefined mention spans or an external mention detection module. It uses a BERT-based biencoder for initial entity ranking followed by a cross-encoder for candidate re-ranking. 
The cross-encoder’s processing, while thorough, is slower compared to the biencoder in ELQ, which can be a disadvantage for applications requiring real-time response, such as conversational systems. Additionally, BLINK’s segmented approach to entity linking, which involves separate processes for mention detection and entity disambiguation, further reduces its suitability for conversational scenarios. 

It is crucial to note that the BLINK model we employ was trained using the Wikipedia KB and has not been exposed to the Fandom KB, ensuring no overlap with the knowledge base used in our evaluations.

\subsection{Knowledge Bases}
\label{subsec:kbs}
We have selected Fandom,%
\footnote{\url{https://www.fandom.com/}}
as the KB for our generalizability analysis. Fandom acts as a hub for `fan-created wikis', covering a range of entertainment topics. 
We use a specific extraction of Fandom for zero-shot EL research called ZESHEL \citep{logeswaran-etal-2019-zero} consisting of 16 Fandom domains and comprising approximately 500,000 entities.   
For our standard setup, we employ the Wikipedia dump from 2019-08-01,%
\footnote{\url{https://github.com/facebookresearch/BLINK/tree/main/elq}} encompassing more than 5 million entities. This version of Wikipedia serves as the standard KB against which ELQ and BLINK are benchmarked.

\subsection{Datasets}

Along with the test set of the zero-shot conversational Reddit dataset introduced in Section \ref{Reddit_data_desc}, we perform experiments using ConEL datasets \citep{joko2021conversational,joko2022personal} and Wikia%
\footnote{\url{https://github.com/lajanugen/zeshel}} documents. The ConEL datasets, comprising ConEL 1 and 2, are derived from various sources and have been specifically curated and human-annotated for the entity linking task against Wikipedia. On the other hand, the Wikia dataset comprises documents featuring mentions of entities from Fandom, with entity annotations contributed by users of the Fandom website. Employing both these datasets allows us to effectively delineate the distinctions between conversational and traditional entity linking which mainly focuses on document-level entity linking.

\subsection{Analysis Scenarios Setups}
\textbf{Generalizability.} ELQ and BLINK share the same entity encoder which is trained on Wikipedia (for language understanding and also for EL) but not on Fandom. To assess their generalizability, the mentioned encoder is used to encode Fandom entities using the first 128 tokens of each entity description.
Our assessment leverages two distinct data sources; our conversational Reddit data and Wikia validation set. 
As BLINK does not support mention detection, we evaluated BLINK's performance in two ways. Once we detected potential mentions using Flair \citep{akbik2018coling} and provided these mentions to BLINK for entity disambiguation. 
Next, to assess BLINK's zero-shot entity disambiguation capabilities, we supply it with gold mention spans of the Wikia validation and Reddit test sets and compare it to a naive baseline (Levenshtein distance).
\begin{table}[ht] 

\centering
\renewcommand{\arraystretch}{1.2} 

\small
\setlength{\tabcolsep}{5pt} 
\begin{tabular}{l cc cc}
\toprule
& \multicolumn{2}{c}{\textbf{Reddit}} & \multicolumn{2}{c}{\textbf{Wikia}} \\
\cmidrule(r){2-3}
\cmidrule{4-5}
& micro & macro & micro & macro \\
\midrule
GT \textbf{+ Edit Distance} & .168 & .161 & .108 & .113 \\
GT \textbf{+ BLINK}        & .288 & .233 & .446  & .457    \\
\bottomrule
\end{tabular}
\caption{Entity disambiguation performance scores given the ground truth mention spans (GT). Evaluation is done on Reddit conversational dataset and on Wikia documents, against Fandom as the knowledge base. Scores are presented as micro-averaged and macro-averaged precision, the first aggregates true positives and false positives across all Fandom domains, while the latter is calculated by averaging domain-specific precision scores.}
\label{table:reddit_results_GT}

\end{table}

\noindent
\textbf{Conversational Context Adapatability.}
This scenario aims to evaluate the EL models' adaptability in a new setting: conversational EL. We evaluate the performance of EL methods in a standard conversational setting using ConEL datasets and Wikipedia as the KB. We then asses the adapatability of ELQ method by fine tuning it on the conversational data (ConEL-2 dataset) and compare its performance with the original mode.

\begin{table}[t]

\centering
\renewcommand{\arraystretch}{1.2} 

\small 
\setlength{\tabcolsep}{3pt} 
\begin{tabular}{@{}l@{~}cc cc cc@{}}
\toprule
 \textbf{} & \multicolumn{2}{c}{\textbf{ConEL1}} & \multicolumn{2}{c}{\textbf{ConEL2-Val}} & \multicolumn{2}{c}{\textbf{ConEL2-Test}} \\ 
\cmidrule(r){2-3}
\cmidrule(r){4-5}
\cmidrule{6-7}
& \scriptsize \textbf{MD} & \scriptsize \textbf{EL} & \scriptsize \textbf{MD} & \scriptsize \textbf{EL} & \scriptsize \textbf{MD} & \scriptsize \textbf{EL} \\
\midrule
 GENRE & .350 & .211 & .290 & .252 & .320 & .299 \\
TagMe & .510 & .375 & .559 & .478 & .611 & .504 \\
 WAT & .416 & .336 & .616 & .539 & .613 & .519 \\
 REL & .462 & .245 & .304 & .244 & .279 & .231 \\
 CREL & \textbf{.559} & .429 & \textbf{.742} & \textbf{.651} & \textbf{.729} & .597 \\
\midrule
 Flair + BLINK & .279 & .166 & .267 & .216 & .257 & .200 \\
 ELQ  & .533 & \textbf{.431} & .596 & .516 & .642 & .575 \\
 ELQ (FT) & .459 & .358 & .706 & .617 & .714 & \textbf{.616} \\
\bottomrule
\end{tabular}
\caption{Entity linking results on ConEL datasets, reported by $F_1$-scores (top rows from \citealp{joko2022personal}). 
Flair+BLINK and ELQ use Wikipedia for both training and inference. ELQ (FT) denotes fine tuning on conversational data (ConEL-2 train set).}
\label{table:linking_results}
\end{table}

\section{Results and Discussion}

\subsection{Are Zero-Shot EL Models Generalizable?} 
We employed Flair+BLINK and ELQ as end-to-end zero-shot entity linking systems evaluating their generalizability on Reddit coversations and Wikia documents. Results in Table \ref{table:reddit_results} reveal a significantly low performance when these systems are tested against Fandom without any pre-training on this specific KB, in both documents and conversations.
This stark underperformance raises questions regarding the practicality and reliability of these systems as zero-shot EL solutions when confronted with novel, domain-specific knowledge bases in the real-world.
The results depict substantial scope for improvement in the mention detection capabilities of both Flair and ELQ. By inspecting the predictions, we realized that numerous text spans are considered as possible correct mentions by Flair/ELQ, many of which do not align with the gold mentions in the Wikia and Reddit datasets. Given that annotations in both datasets is done by users, this raises the question of whether these methods can model entity saliency so that predictions are relevant and align with the user expectations.
Considering table \ref{table:reddit_results_GT} we observe that even given the gold mention spans, correctly linking entities in conversations is more challenging for BLINK than in documents, highlighting the complexity of this environment. This highlights the need for better entity disambiguation techniques that consider and leverage conversational characteristics for improved disambiguation.

\subsection{Are Zero-Shot EL Models Adaptable to Conversational EL Task?}
We analyzed adaptability of end-to-end EL systems, specifically Flair+BLINK and ELQ, for disambiguating entity mentions in conversations without prior training in this context—a zero-shot setup. 
Findings are summarized in Table \ref{table:linking_results}, where the top rows show common EL systems evaluated by \citealp{joko2022personal}, with only CREL~\cite{joko2022personal} being optimized for conversations and the rest (GENRE, TagMe, WAT and REL) are general-purpose EL systems. Results for Flair+BLINK and ELQ can be found in the second part of table.
Flair underperforms in conversation mention detection, while fine tuned ELQ (adapted to conversational setup) excels in both mention detection and entity disambiguation, outdoing most models except CREL which is optimized for conversations. The adaptability of ELQ is likely due to the end-to-end MD and ED training, as well as  similarity to the domain it is initially trained on.

\section{Conclusions and Future Work}


This study re-examined the efficacy of current EL models in conversational scenarios with limited data and KB resources. Motivated by the real-world challenges frequent when integrating EL components into conversational assistants, we recognized overlooked practical limitations in zero-shot EL research.  We showed that current zero-shot EL models critically underperform when introduced to a new KB at inference time, due to shortcomings in both mention detection and entity disambiguation functions. These results highlight the need for designing better end-to-end zero-shot EL systems that are reliable in various tasks and KB constraint scenarios. We conclude that the evaluation approaches being used so far in EL literature to evaluate zero-shot EL models are quite naive and not representative of the user's perspective on entity saliency, a crucial point when in interactive systems. For future work, we will leverage our curated dataset to advance model capabilities.


\section{Limitations}
Our experiment setup involves the use of a new KB, however, the number of EL systems allowing such a use case is very limited. On the other hand, end-to-end EL systems capable of integrating mention detection and entity disambiguation is also limited. These made our choice of models to evaluate quite restricted. Additionally, to test the capabilities of models in zero-shot conversational setup, we needed a conversational dataset that is annotated by entities in a specific-domain KB with long-tail entities. Such data is usually proprietary and not open-access, thereby we had to simulate such a scenario. It would be interesting to assess whether our results hold for other domains.

\section{Acknowledgement}
This publication is part of the project LESSEN with project number NWA.1389.20.183 of the research program NWA ORC 2020/21 which is (partly) financed by the Dutch Research Council (NWO). 

We also would like to express our gratitude towards Emma Gerritse and Hideaki Joko for their insightful advice and inspiring discussions, which influenced the direction of the paper.

 \bibliography{custom}

\appendix
\label{sec:appendix}

\section{Replicating BLINK Results on Fandom}
To ensure our results are comparable to those reported in \citep{wu-etal-2020-scalable}, we used their Wikipedia-trained bi-encoder and cross-encoder model (the only trained models they released) and evaluated it on Wikia's validation set using the evaluation approaches and metrics employed by BLINK's authors. We included the results in Table \ref{tab:wikia_validation_performance}. As this model is only trained on Wikipedia and the scores in BLINK paper are based on a Fandom-trained model, the performance is close but still lower than the ones reported by the authors. 

\begin{table}[h]

\centering
\begin{tabular}{l@{~}cccc}
\toprule
\textbf{Dataset} & \textbf{R@64} & \textbf{Bi} &  \textbf{Cross} & \textbf{All} \\ 
\midrule
Elder Scrolls & .896 &.354 &  .4722 & .423 \\
Muppets & .819 &.511 &  .650 & .533 \\
Ice Hockey & .857 &.453 & .484 & .415 \\
Coronation Street &.698 & .208 &  .632 & .442 \\ 
\midrule
\textbf{Macro average} & \textbf{.818} & \textbf{.382} & \textbf{.560} & \textbf{.453} \\
\bottomrule
\end{tabular}
\caption{Performance of BLINK on Wikia Validation Set. \emph{R@64, Bi, Cross,} and \emph{All} represent Biencoder Recall@64, Biencoder accuracy, Crossencoder normalized accuracy, and overall unnormalized accuracy, respectively. The scores reported align with the evaluation approach used in BLINK.}
\label{tab:wikia_validation_performance}
\end{table}

\section{Evaluation Metrics}
We evaluate the performance of the EL systems across three aspects; mention detection (MD), entity disambiguation (ED) \citep{cornolti2013framework}, and entity linking (EL). To assess mention detection (MD) we employ a strict matching criterion, where a predicted span is deemed accurate only if it has complete overlap with the corresponding gold standard mention span. Given the entity catalogue $E$, let $T$ and $\hat{T}$ be the set of gold and predicted mention and entity pairs respectively. Consequently, with our matching criterion, the set of final true positives for entity linking will be defined as: 
\begin{align*}
    C =\{  e \in E \mid [m_s, m_e] = [\hat{m}_s, \hat{m}_e], \\ 
     (e, [m_s, m_e]) \in T, (e, [\hat{m}_s, \hat{m}_e]) \in \hat{T} \}
\end{align*}
%
%
We report precision ($p$), recall ($r$) and F1-score ($F_1$) for the three aspects whenever it is relevant. For generalizability experiments, both micro and macro averaging are used to report the scores across multiple Fandom domains.

\section{Zero-Shot Conversational EL Reddit Data}
\label{sec:data_split}
Our final threads timeline spans from April 27, 2010, to October 31, 2018. Threads dated up to January 1, 2015, were allocated to the training set. For the test set, we selected the densest thread from conversations post-January 1, 2015, as the test thread, incorporating the rest into the training set. We include samples of the dataset in Table~\ref{tab:reddit_samples}.


\clearpage
\onecolumn
\afterpage{
\begin{longtable}{@{} p{3cm} p{13cm} @{}}
    \toprule
    \multicolumn{2}{c}{\textbf{Conversation Sample \#1}} \\
    \midrule
    \textbf{Utterances} \\
    \multicolumn{2}{p{16cm}}{
        User \#1: \textit{I know this is a far shot but I had an idea today that I thought I would share .} \newline
        User \#2: \textit{What if BB 8 is actually Boba Fett ? BoBa Fett , BB Eight .} 
        \newline
        User \#3: \textit{His head is inside of BB 8!} 
        \newline
        User \#4: \textit{Exactly ! BB 8 confirmed as an updated \textcolor{blue}{BT 16 droid} : The B'omarr Monks used these droids to carry brains of those who had achieved enlightenment .} 
        \newline
        User \#5: \textit{OMG THAT IS NOT CANON ANYMORE} 
    }
    \\
    \multicolumn{2}{l}{\textbf{Mentions and Entities}} \\    
    \multicolumn{2}{p{16cm}}{
        \textbf{Mention \#1}:  BT 16 droid 
        \newline
        \textbf{Entity \#1}:  \url{https://starwars.fandom.com/wiki/BT-16_perimeter_droid}
    } 
    \\
    \midrule
    \multicolumn{2}{c}{\textbf{Conversation Sample \#2}} \\
    \midrule
    \textbf{Utterances} \\
    \multicolumn{2}{p{16cm}}{
        User \#1: \textit{Deck building Hey Guys is the structure deck saga of the Blue eyes White Dragon Still viable to start with ? And is it worth to buy that structure deck 3 times ? Thanks for helping} 
        \newline
        User \#2: \textit{Nope . Get Structure Deck : Seto Kaiba instead} 
        \newline
        User \#3: \textit{Not even for casual Play ?} 
        \newline
        User \#4: \textit{In that case get the Legendary Dragon Decks or Legendary Decks 2. Otherwise buy them singles instead} 
        \newline
        User \#5: \textit{\textcolor{blue}{deck} you mean This One ?} 
    } \\
    \multicolumn{2}{l}{\textbf{Mentions and Entities}} \\
    \multicolumn{2}{p{16cm}}{
        \textbf{Mention \#1}:  deck
        \newline
        \textbf{Entity \#1}:  \url{https://yugioh.fandom.com/wiki/Legendary_Dragon_Decks}
    } \\
    \midrule   
    \multicolumn{2}{c}{\textbf{Conversation Sample \#3}} \\
    \midrule
    \textbf{Utterances} \\
    \multicolumn{2}{p{16cm}}{
        User \#1: \textit{Should Konami release a small Link monster set Like the title says should Konami release maybe a 30 card set for just Link monsters when they drop that was my biggest complaint about synchros and XYZ that they didn't release a small set of just those card type that was mostly filled with generic monsters to help build the extra deck with .} \newline
        User \#2: \textit{They could release a links starter deck like they did \textcolor{blue}{for Synchros}.}
        \newline
        User \#3: \textit{actually \textcolor{blue}{they did} but it's garbage} 
        \newline
        User \#4: \textit{I think it was good for learning how to synchro before they came out in a set . Should do the same for links .}      
        \newline
        User \#5: \textit{Again , a \textcolor{blue}{link strater deck} already exists . The problem is that it's crap .} 
    } \\
    \multicolumn{2}{l}{\textbf{Mentions and Entities}} \\
    \multicolumn{2}{p{16cm}}{
        \textbf{Mention \#1}: for Synchros
        \newline
        \textbf{Entity \#1}:  \url{https://yugioh.fandom.com/wiki/The_Duelist_Genesis}
        \newline
        \textbf{Mention \#2}: they did
        \newline
        \textbf{Entity \#2}:  \href{https://yugioh.fandom.com/wiki/Starter_Deck:_Yu-Gi-Oh!_5D\%27s}{https://yugioh.fandom.com/wiki/Starter\_Deck:\_Yu-Gi-Oh!\_5D\%27s}       
        \newline
        \textbf{Mention \#3}: link strater deck
        \newline
        \textbf{Entity \#3}: \url{https://yugioh.fandom.com/wiki/Starter_Deck_2017} 
    } \\
    \\[3.2cm]
    \midrule
        \multicolumn{2}{c}{\textbf{Conversation Sample \#4}} \\
    \midrule
    \textbf{Utterances} \\
    \multicolumn{2}{p{16cm}}{
        User \#1: \textit{Secret New Hero : Jabba ? Jabba as a hero anyone ?} 
        \newline
        User \#2: \textit{Was there actually ever a Hutt Jedi ? What about Tusken Raider ? Imagine Jabba being a bullet sponge with ATAT health .} 
        \newline
        User \#3: \textit{\textcolor{blue}{Beldorian the Hutt} was a Jedi , but fell to the darkside . \textcolor{blue}{Sharad Hett} was a Jedi who left the Order and joined the Tuskens . His son \textcolor{blue}{A'Sharad Hett} eventually became Darth Krayt .} 
        \newline
        User \#4: \textit{\textcolor{blue}{Beldorian the Hutt} was killed in a light saber duel by Leia Organa Solo . What .} 
        \newline
        User \#5: \textit{Leia becomes a jedi in the old EU} 
    } \\
    \multicolumn{2}{l}{\textbf{Mentions and Entities}} \\
    \multicolumn{2}{p{16cm}}{
        \textbf{Mention \#1}: Beldorian the Hutt
        \newline
        \textbf{Entity \#1}:  \url{https://starwars.fandom.com/wiki/Beldorion}
        \newline
        \textbf{Mention \#2}: Sharad Hett
        \newline
        \textbf{Entity \#2}: \url{https://starwars.fandom.com/wiki/Sharad_Hett}
        \newline
        \textbf{Mention \#3}: A'Sharad Hett
        \newline
        \textbf{Entity \#3}: \url{ https://starwars.fandom.com/wiki/Darth_Krayt}
        \newline
        \textbf{Mention \#4}: Beldorian the Hutt
        \newline
        \textbf{Entity \#4}:  \url{https://starwars.fandom.com/wiki/Beldorion}        
    } \\
    \midrule
    \multicolumn{2}{c}{\textbf{Conversation Sample \#5}} \\
    \midrule
    \textbf{Utterances} \\
    \multicolumn{2}{p{16cm}}{
        User \#1: \textit{[Question ?] Borreload Dragon and Eater of Millions Whats the interaction between these two cause bouth trigger at the begin of the damage step} \newline
        User \#2: \textit{Both cards trigger at the same time . Turn player's trigger is added to the chain first according to SEGOC . CL1: Borreload CL2: Eater Resolve the chain backwards : Eater banishes Borreload Borreload resolves without effect , as it no longer points at any zones} 
        \newline 
        User \#3: \textit{What about the other way around ? I . e . if the turn player controls Eater . CL1: Eater CL2: Borreload Resolve backwards : Borreload takes control of Eater Eater banishes Borreload ?} 
        \newline
        User \#4: \textit{Borreload can only trigger when it's attacking , so it's controlled by the turn player . This means Borreload will be CL1 under \textcolor{blue}{SEGOC} , the turn player's optional effects Borreload come before the non turn player's optional effects Eater .} 
        \newline
        User \#5: \textit{Ahhhh I didn't realise it only triggered when attacking . Thanks !} 
    } \\
    \multicolumn{2}{l}{\textbf{Mentions and Entities}} \\
    \multicolumn{2}{p{16cm}}{
        \textbf{Mention \#1}:  SEGOC 
        \newline
        \textbf{Entity \#1}:  \url{https://yugioh.fandom.com/wiki/Simultaneous_Effects_Go_On_Chain}
    } 
    \\
    \bottomrule
    \caption{Sample conversations from the dataset.} 
    \label{tab:reddit_samples}     
\end{longtable}
}

\end{document}